%% file: 0-Main.tex
\documentclass[sigconf]{acmart}

\usepackage{enumitem}
\usepackage{algorithm}
\usepackage{bbm}
\usepackage{dsfont}
\usepackage{amsmath}

\usepackage{amssymb}
\usepackage{multirow}

\AtBeginDocument{%
  }

\setcopyright{acmlicensed}
\copyrightyear{2024}
\acmYear{2024}
\setcopyright{acmlicensed}
\acmConference[CIKM '24] {Proceedings of the 33rd ACM International Conference on Information and Knowledge Management}{October 21--25, 2024}{Boise, ID, USA.}
\acmBooktitle{Proceedings of the 33rd ACM International Conference on Information and Knowledge Management (CIKM '24), October 21--25, 2024, Boise, ID, USA}
\acmDOI{10.1145/3627673.3679900}
\acmISBN{979-8-4007-0436-9/24/10}
\begin{document}

\title{Does Knowledge Localization Hold True? 
Surprising Differences Between Entity and Relation Perspectives in Language Models}

\author{Yifan Wei}
\authornote{Both authors contributed equally to this research.}
\orcid{0000-0002-8870-7304}
\author{Xiaoyan Yu}
\orcid{0009-0006-2314-7867}
\authornotemark[1]
\affiliation{%
\institution{$\text{C}^\text{2}$DL, Institute of Automation, Chinese Academy of Sciences
}
\city{Beijing}
\country{China}
}
\email{weiyifan2021@ia.ac.cn}
\email{xiaoyan.yu@bit.edu.cn}

\author{Yixuan Weng}
\orcid{0000-0002-9720-8689}
\affiliation{%
  \institution{$\text{C}^\text{2}$DL, Institute of Automation, Chinese Academy of Sciences}
  \city{Beijing}
   \country{China}
  }
\email{wengsyx@gmail.com}

\author{Huanhuan Ma}
\orcid{0000-0002-7151-9550}
\affiliation{%
  \institution{NLPR, Institute of Automation, Chinese Academy of Sciences}
  \city{Beijing}
   \country{China}
 }
\email{mahuanhuan2021@ia.ac.cn}

\author{Yuanzhe Zhang}
\authornote{Corresponding authors.}
\orcid{0000-0001-9905-9501}
\affiliation{%
  \institution{$\text{C}^\text{2}$DL, Institute of Automation, Chinese Academy of Sciences}
  \city{Beijing}
   \country{China}
  }
\email{yzzhang@nlpr.ia.ac.cn}

\author{Jun Zhao}
\orcid{0000-0003-3370-2263}
\affiliation{%
\institution{$\text{C}^\text{2}$DL, Institute of Automation, Chinese Academy of Sciences}
   \city{Beijing}
   \country{China}
  }
\email{jzhao@nlpr.ia.ac.cn}

\author{Kang Liu}
\authornotemark[2]
\orcid{0000-0002-6083-8433}
\affiliation{%
\institution{$\text{C}^\text{2}$DL, Institute of Automation, Chinese Academy of Sciences}
\institution{Shanghai Artificial Intelligence Laboratory}
  \city{Beijing}
   \country{China}
  }
\email{kliu@nlpr.ia.ac.cn}

\renewcommand{\shortauthors}{Yifan Wei et al.}

\begin{abstract}

Large language models  encapsulate knowledge and have demonstrated superior performance on various natural language processing tasks.
Recent studies have localized this knowledge to specific model parameters, such as the MLP weights in intermediate layers. 
This study investigates the differences between entity and relational knowledge through knowledge editing. 
Our findings reveal that entity and relational knowledge cannot be directly transferred or mapped to each other. This result is unexpected, as logically, modifying the entity or the relation within the same knowledge triplet should yield equivalent outcomes. 
To further elucidate the differences between entity and relational knowledge, we employ causal analysis to investigate how relational knowledge is stored in pre-trained models. Contrary to prior research suggesting that knowledge is stored in MLP weights, our experiments demonstrate that relational knowledge is also significantly encoded in attention modules. This insight highlights the multifaceted nature of knowledge storage in language models, underscoring the complexity of manipulating specific types of knowledge within these models.

\end{abstract}



\maketitle

\input{1-Introduction}

\input{5-RelatedWork}

\input{2-Preliminary}
\input{3-Method}
\input{4-Experiments}
\input{6-Conclusion}




\newpage
\bibliographystyle{ACM-Reference-Format}
\bibliography{ref}



\end{document}

%% file: 1-Introduction.tex
\section{Introduction}
    

    Large language models (LLMs), trained on extensive knowledge corpora such as Wikipedia, encapsulate a vast amount of factual knowledge and demonstrate exceptional performance in various natural language tasks. 
    Consequently, LLMs are often regarded as knowledge bases that underpin knowledge-oriented tasks \cite{yu2021multi, geva2021transformer, xia2022medconqa, wei2023menatqa, ma2023ex, yu2024neeko, li2024relational}. 
    However, leveraging the knowledge within these models effectively requires understanding the mechanisms by which LLMs store and manage factual knowledge. 
    This understanding is crucial for tasks such as model editing \cite{dai2022knowledge,meng2022locating,meng2022mass, chen2024journey, chen2024knowledge}, which involves modifying the knowledge embedded in the models.
    
    \begin{figure}[htbp]
   \centering
   \includegraphics[width=1\linewidth]{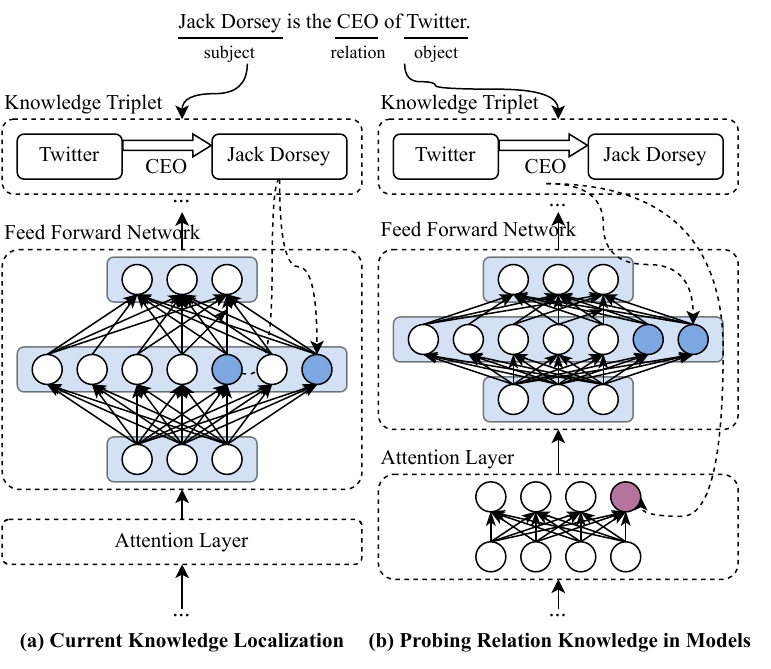}
   \caption{Knowledge stored within model parameters.}
   \label{fig:example}
   \end{figure}


    Current studies \cite{petroni2019language, meng2022locating, dai2022knowledge, hu2024wilke, chen2024knowledge} have focused on studying the knowledge embedded in LLMs. 
    These works have considered knowledge in the form of triplets $(s, r, o)$, which include the head entity (subject, $s$), tail entity (object, $o$), and their relation $r$, as shown in Figure \ref{fig:example}. 
    They have examined how language models encapsulate knowledge in their parameters. 
    For instance, \citet{dai2022knowledge} employed a knowledge attribution method and identified specific neurons that express factual knowledge, while \citet{meng2022locating} used causal tracing to find strong causality between subjects and the MLP module. 
    However, these studies primarily investigate the knowledge in LLMs from the entity perspective. The total different observations could be conducted if we address the same knowledge from the relation.
    Theoretically, a piece of knowledge includes both entities and their relations; without either, it is incomplete. 
    Therefore, entities and relations are supposed to be equivalent in this context, a premise upon which much current work in model editing is based, given the need to modify knowledge in the model parameters.

    Nevertheless, current studies have not yet explored whether such equivalence stands. 
    To fill this gap, we investigate the differences between entity and relation in this paper. 
    To explore this potential equivalence, we employ model editing, a technique for updating or correcting new or erroneous knowledge in language models. 
    We aim to determine whether changes yield consistent outcomes by modifying entity or relational knowledge, observing the effects from both perspectives. 
    Ideally, the effects should be identical since the edited knowledge pertains to the same piece.
    To further elucidate the differences in where relational and entity knowledge is stored, we examine how relational knowledge is stored in auto-regressive transformer models. 
    We employ causal analysis to explore the relationship between relational knowledge and the various modules of LLMs. 
    Our probing leads to two surprising conclusions: 
        (1) factual knowledge is not stored as a single unit; relations and entities are represented separately within the model parameters, as simply illustrate in Figure \ref{fig:example}(b); 
        (2) editing from entity and relational perspectives does not yield the same outcomes, which means the previous located knowledge neurons in previous work are questionable.

    The findings in this work have profound implications for understanding and utilising LLMs in knowledge representation and model editing.
    This revelation challenges the validity of existing evaluation methods that assess the success of model edits based on this flawed assumption of equivalence. 
    By revealing these discrepancies, our work provides a new foundation for future research and development in LLM-related tasks, such as model editing.

%% file: 5-RelatedWork.tex
\section{Related Work} \label{sec:relatedwork}

    As factual information continues to evolve, the knowledge stored in large language models (LLMs) can become outdated or incorrect. Hence, there is an urgent need to facilitate timely updates of inappropriate knowledge in LLMs while preserving other valuable knowledge. Recently, this issue has garnered significant attention from researchers. Certainly, both parameter-efficient fine-tuning and incremental learning techniques provide avenues for modifying LLMs. However, it is essential to note that these approaches may be prone to overfitting and can incur substantial computational costs, especially when applied to LLMs with an extremely large parameter scale. 
    To address these issues, \citet{sinitsin2019editable} proposes Model Editing, which aims to efficiently and accurately alter the factual knowledge stored within models. Presently, there are three primary types of model editing approaches:
    1) Memory-based Method: These techniques utilize additional trainable parameters to store memory or learn the required adjustments ($\Delta$) for knowledge updating in the LLMs \citep{de2021editing, mitchell2021fast, mitchell2022memory, dong2022calibrating, huang2022transformer}.
    2) Locate-Then-Edit Method: These approaches employ causal mediation analysis to locate knowledge neurons in LLMs and subsequently modify these recognized regions \citep{dai2022knowledge, meng2022locating, meng2022mass}. This paper primarily explores this knowledge localization method.
    3) In-Context Knowledge Editing Method: These methods are a training-free paradigm where knowledge editing is achieved directly by concatenating demonstrations within the input context \citep{zheng2023can, zhong2023mquake}. 
    This paper primarily explores the second type, the Locate-Then-Edit method.

%% file: 3-Method.tex
\section{Background \& Methodology}\label{sec:method}

    
    \subsection{Task Definition} \label{sec:task_def}
        Assume that knowledge $\mathcal{K} = \{x,y\}$ is stored in language model in the form of triples $(s,r,o)$.
        The objective of model editing is to modify a base model $f_{\theta}$, parameterized by $\theta$, which maps the text prompt $P$ as input $x$ to gain control over the model's prediction outputs $y$, expressed as:
        \begin{equation}\label{equation1}
            f_{\theta}(x)=\underset{\theta}{\operatorname{argmax}} p_\theta(y \mid P).
        \end{equation}
        To modify the prediction results, model editing aims to update the model parameter $\theta^*$ with $f(x;\theta^*) = y^* $. 
        Editing reliability is needed to change prediction from $y$ to $y^*$.

        \subsection{Model Editing Methods}\label{sec:KE_methods}
            To explore the connection between model parameters and knowledge, we apply model editing techniques to modify the parameters of transformer-based language models. 
            In this section, we describe the model editing methods applied.
            
            To modify specific knowledge $\mathcal{K}$ in a model, we adjust the model weight parameters $W$ associated with $\mathcal{K}$. 
            The objective is to optimize the hidden states of both the Attention and MLP components.
            The target weight $\hat{W}$ is defined as:
            \begin{equation}
                \hat{W} \triangleq \underset{W}{\operatorname{argmin}}\left(\sum_{i=1}^n\left\|W k_i-v_i\right\|^2+\sum_{i=n+1}^{n+u}\left\|W k_i-v_i\right\|^2\right),
            \end{equation}
            where $k_i$ represent the knowledge index vector obtained through the $i$-th prompt $x_i$ and $v_i$ represent the target knowledge representation. 
            $\sum_{i=1}^{n} \left\|W k_i - v_i \right\|^2$ indicates the retention of $n$ pieces of knowledge, and $\sum_{i=n+1}^{n+u} \left\|W k_i - v_i \right\|^2$ indicates the modification of $u$ pieces of knowledge.
            We compute a target vector $v_i$ to replace the original hidden state $h_i^L$ by optimizing the residual vector $\delta_i$ using gradient descent:
            \begin{equation}
                v_i=h_i^L+\delta_i=h_i^L+\underset{\delta_i}{\operatorname{argmin}} \frac{1}{N} \sum_{j=1}^N-\log \mathbb{P}_{\theta\left(h_i^L+=\delta_i\right)}\left[y_i \mid x_i \right].
            \end{equation}
            Given prompt $x_i$ to update knowledge $\mathcal{K}$, we optimize $\delta_i$ to maximize the model’s prediction of the desired output $y_i$.

        \begin{figure*}[htbp]
            \centering
            \includegraphics[width=1\linewidth,height=0.18\textheight]{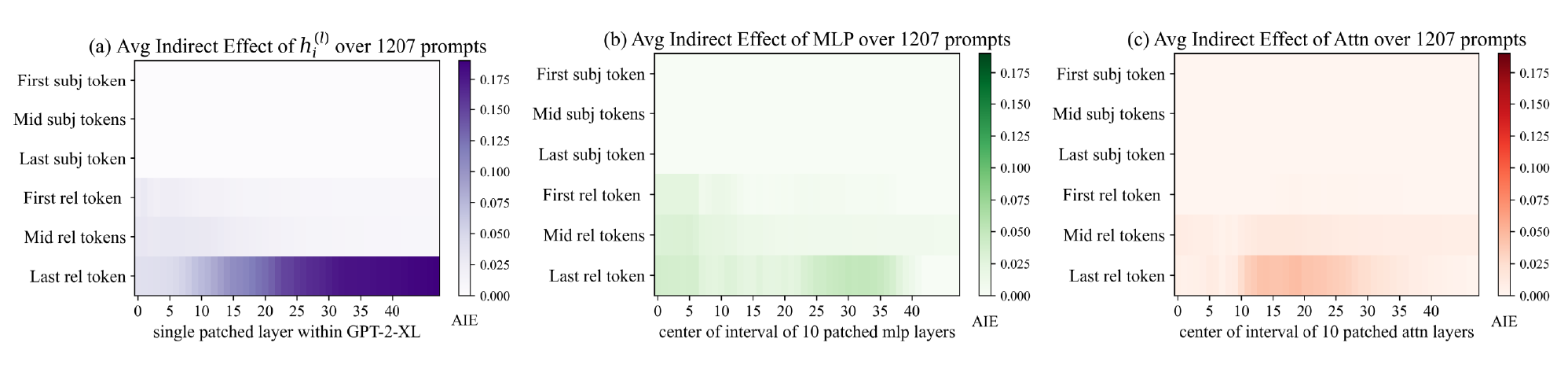}
            \caption{Causal tracing results of individual model components. }
            \label{fig:AIE}
        \end{figure*}

        \subsection{Locating Relation Knowledge}\label{sec:relation_loc}
            \paragraph{\textbf{Casual Tracing}}
            To locate the relations $r$ within factual triplets $(s, r, o)$ in model parameters, we analyze and identify the knowledge neurons with the strongest causal effect on these relations. 
            We employ causal tracing for this purpose, following this procedure:
            \begin{description}[leftmargin=10pt]
                \item[\textbf{Step 1}] \textbf{Clean run.} 
                A factual prompt $x$ is passed into the model $f_{\theta}$ and collect all hidden activations $\{h_i^{(l)}\ |\ i \in [1,T], l \in [1,L]\}$, where $T$ is number of input tokens in $x$ and $L$ is number of layers.
                \item[\textbf{Step 2}] \textbf{Corrupted run.} 
                The relation embeddings $[h_1^{(0)}, h_2^{(0)}, \linebreak \ldots, h_T^{(0)}]$ are obfuscated by adding a term $\epsilon$ to each $h_i^{(0)}$, where $\epsilon \sim \mathcal{N}(0, \nu)$ and $\nu$ is set to three times the empirical standard deviation of the embeddings. 
                This results in a set of corrupted activations $\{h_{i \ast}^{(l)} \mid i \in [1, T], l \in [1, L]\}$.
                \item[\textbf{Step 3}] \textbf{Corrupted-with-restoration run.}
                The model $f_{\theta}$ perform computations on the noisy embeddings, as in the corrupted baseline. 
                However, at a specific token $\hat{i}$ and layer $\hat{l}$, $f_{\theta}$ is intervened to output the clean state $h^{(l)}_{\hat{i}}$. 
                After this point, all subsequent computations proceed without further intervention.
            \end{description}
            $\mathbb{P}[y]$, $\mathbb{P_*}[y]$, and $\mathbb{P}_{*, \operatorname{clean} h_i^{(l)}}[y]$ is defined as the probability of final prediction $y$ under the clean, corrupted, and corrupted-with-restoration runs, respectively. 
            The indirect effect (IE) of a particular hidden state $h_i^l$ is calculated as:
            \begin{equation}
                \mathrm{IE}=\mathbb{P}_{*, \operatorname{clean} h_i^{(l)}}[y]-\mathbb{P}_*[y].
            \end{equation}
        
            \paragraph{\textbf{Severed Causal Analysis}}
            To gain a clearer understanding of the impact of MLP and Attention layers, we perform severed causal tracing analysis using a modified causal graph, following \cite{meng2022locating}. 
            In the corrupted-with-restoration-run, we freeze the MLP and Attention modules to the corrupted run value so that it’s unaffected by the inserting of clean state $h_i^{(l)}$. 
            This can viewed as severing the MLP and Attention computations from the original computation graph. 
            The propagation of noise in the model follows:
            \begin{equation}
                \begin{aligned}
                    h^{(l)}_i &= h^{(l-1)}_i + \text{sever}(a^{(l)}_i , m^{(l)}_i) \\
                    a^{(l)}_i &= \text{attn}^{(l)}\left(h_{1}^{(l-1)}, h_{2}^{(l-1)}, \dots, h_{i}^{(l-1)}\right) \\
                    m^{(l)}_i &= W^{(l)}_{proj} \sigma \left( W^{(l)}_{fc} \gamma \left( a^{(l)}_i + h^{(l-1)}_i \right) \right),
                \end{aligned}
            \end{equation}
            where the function $\text{sever}(\cdot)$ denotes the server operation, which separates the MLP or Attention computations from the  model.

%% file: 4-Experiments.tex
\section{Experiments}\label{sec:exp}
   To investigate how knowledge is stored within model parameters, we outline the following Research Questions (RQs):
    \begin{itemize}[leftmargin=15pt]
        \item \textbf{RQ1}: Where is relational knowledge stored? Is it stored in the same manner as entity knowledge within MLPs?
        \item \textbf{RQ2}: Are relation and entity knowledge equally significant in knowledge triplets, regardless of their storage location?
    \end{itemize}

    \subsection{Experimental Setups}

            
            In the experiments, we use GPT-2 XL (1.5B) and GPT-J (6B) as the base language models. 
            The experiments are conducted with four NVIDIA RTX A6000 GPUs and ten NVIDIA GeForce RTX 3090 GPUs.
            The evaluation metrics includes Reliability and Generality.
            \textbf{Reliability} quantifies the reliability of the editing process, with higher reliability indicating greater success in editing. 
            To measure reliability, we assess the editing accuracy as follows:
            \begin{equation}
            \mathcal{M}_{r e l}=\mathbb{E}_{\left(x, y^*\right) \sim \mathcal{D}} \left[\mathds{1}_{f\left(x ; \theta^*\left(x, y^*\right)\right)=y^*}\right],
            \end{equation}
            \textbf{Generality} measures the generalization ability of the edited model's predictions across various inputs or contexts.
             \begin{equation}
            \mathcal{M}_{gen}=\mathbb{E}_{\left(\tilde{x}\right) \sim \mathcal{N}(x)} \left[\mathds{1}_{f\left(\tilde{x} ; \theta^*\right)=f\left(x; \theta^* \right)= y^{*} }\right],
            \end{equation}
            where $\tilde{x}$ refers to the rephrased text prompt , $\mathcal{N}(x)$ denotes a set of rephrased prompts equivalent to  $x$.



    \subsection{RQ1: Causal Analysis for Relation}\label{sec:cau_trace}
        We conducted causal tracing analysis to determine the location of relational knowledge within model parameters, with the results illustrated in Figure \ref{fig:AIE}. 
        The procedure of causal tracing analysis is outlined in Section \ref{sec:relation_loc}.
        By varying the mediator across different positions within the prompt and different model components (such as individual states, MLP layers, and attention layers), we calculated the average indirect effect (AIE) across 1207 factual statements.
        The results show that, consistent with prior findings \cite{meng2022locating, meng2022mass}, there is a high AIE score in the last layers of the final token. 
        This indicates that restoring the hidden states of the MLPs in these layers recovers most of the necessary information. 
        Additionally, we observed a high AIE score in the earlier layers for the intentionally corrupted relation tokens, underscoring the importance of these early layers in predicting plausibility.

\begin{figure}[htbp]
            \centering
            \includegraphics[width=0.9\linewidth]{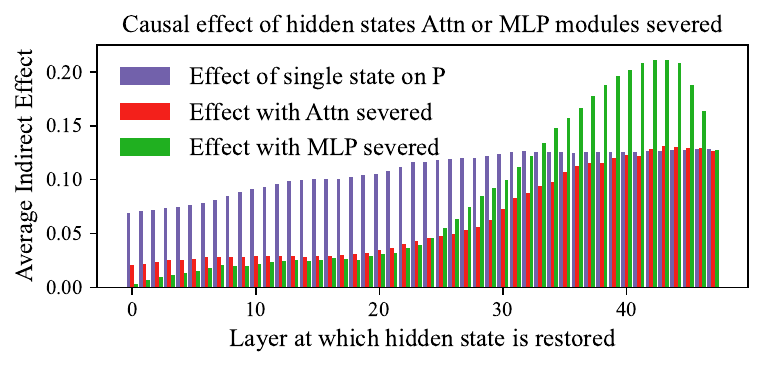}
            \caption{Causal effects by isolating various modules.}
            \label{fig:AIE_for_each_module}
        \end{figure}

        
        Similarly, we noted a pronounced AIE in the middle attention layers of the last corrupted token. 
        We found that the knowledge storage location identified by the relation $r$ in the knowledge triples is strongly correlated with both MLP layers and attention layers, as shown in Figure \ref{fig:AIE_for_each_module}. 
        This conclusion differs from previous works identifying knowledge storage in lower MLP layers via entity localization. 
        We discover that knowledge expression localized through relations is closely associated with higher MLP layers and mid-to-upper attention layers. 
        When exploring model knowledge expression from an entity perspective to a relation perspective, the causal locations of knowledge expression in the model change significantly. 
        This indicates that the storage location of knowledge in the model parameters is complex and cannot be simply determined by causal tracing from a single perspective, assuming knowledge is isolated in specific model layers.
        Therefore, we believe that modifying the corresponding model parameters to control the expression of knowledge through such localization is unreasonable.

    \subsection{RQ2: Probing the Equivalence }\label{sec:ke_e&r}
        Under the assumption that entity and relation perspectives are logically equivalent in knowledge triplets, as illustrated in Figure \ref{fig:AIE_for_each_module}, entity knowledge and relational knowledge are considered interchangeable. 
        Based on this assumption, we hypothesize that modifying entity knowledge by altering relational knowledge is theoretically possible. 
        To validate this hypothesis, we apply model editing techniques to modify knowledge in language models from relational and entity perspectives and observe whether the effects remain the same.
        Table \ref{tab:edit_rel} presents the evaluation results from both relation and entity perspectives after applying relation-based model editing methods. 
        Contrary to our assumption, we are surprised the evaluation score for entity lags far from that for relation. 
        Editing relation knowledge achieves high metrics for relation, indicating that these editing methods are effective. 
        However, the results for entity knowledge are noticeably lower, suggesting that editing relation does not effectively alter entity knowledge.
        This is puzzling because entities and relations within the same triplet define a piece of knowledge. 
        Altering any part of the triplet should theoretically alter the entire triplet, implying equivalence.

\begin{table}[htbp]
    \centering
    \caption{Performance with edited relation knowledge.}
    \resizebox{0.8\columnwidth}{!}{%
    \begin{tabular}{lcccccc}
        \toprule
        \multirow{3}{*}{\textbf{Method}} & \multicolumn{4}{c}{\textbf{Entity Knowledge}} & \multicolumn{2}{c}{\textbf{Relation Knowledge}} \\
        \cmidrule(lr){2-5} \cmidrule(lr){6-7}
        & \multicolumn{2}{c}{\textbf{Reliability}} & \multicolumn{2}{c}{\textbf{Generality}} & \textbf{Reliability} & \textbf{Generality} \\
        \midrule
        \multicolumn{7}{c}{\textbf{GPT-2 XL}} \\
        \midrule
        FT   & 23.92 &        & 25.44 &        & 98.79 & 79.03 \\
        KN   & 22.53 &        & 24.61 &        & 97.52 & 76.16 \\
        MEND & 22.33 &        & 24.63 &        & 100.0 & 83.24 \\
        ROME & 27.92 &        & 28.12 &        & 99.99 & 84.47 \\
        MEMIT& 24.15 &        & 24.63 &        & 91.36 & 76.24 \\
        \midrule
        \multicolumn{7}{c}{\textbf{GPT-J}} \\
        \midrule
        MEND & 15.51 &        & 17.99 &        & 100.0 & 81.52 \\
        ROME & 30.95 &        & 31.87 &        & 100.0 & 95.97 \\
        MEMIT& 18.92 &        & 19.37 &        & 100.0 & 88.50 \\
        \bottomrule
    \end{tabular}
    }
    \label{tab:edit_rel}
\end{table}

Table \ref{tab:edit_ent} presents the 
evaluation results from relation and entity perspectives after applying entity-based editing methods \cite{meng2022locating, meng2022mass}. 
        The results in Table \ref{tab:edit_ent} show that the evaluation results are relatively stable with rather minimal fluctuation. 
        The reliability of relation knowledge has improved, but there is a significant decrease in the generality metrics. 
        These findings suggest that model editing from an entity perspective can potentially alter the relation information between pieces of knowledge. Howerver, changes are inconsistent.
        \textbf{The above findings indicate that editing entity knowledge and relation knowledge are not exactly equivalent.}
        
        \begin{table}[htbp]
        \centering
        \caption{The performance by editing entity knowledge.}
        \resizebox{0.8\columnwidth}{!}{%
        \begin{tabular}{lcccccc}
            \toprule
            \multirow{3}{*}{\textbf{Method}} & \multicolumn{4}{c}{\textbf{Entity Knowledge}} & \multicolumn{2}{c}{\textbf{Relation Knowledge}} \\
            \cmidrule(lr){2-5} \cmidrule(lr){6-7}
            & \multicolumn{2}{c}{\textbf{Reliability}} & \multicolumn{2}{c}{\textbf{Generality}} & \textbf{Reliability} & \textbf{Generality} \\
            \midrule
            \multicolumn{7}{c}{\textbf{GPT-2 XL}} \\
            \midrule
            ROME & 99.93 &        & 96.6 &        & 96.12 & 74.46 \\
            MEMIT& 93.88 &        & 79.6 &        & 97.28 & 76.01 \\
            \midrule
            \multicolumn{7}{c}{\textbf{GPT-J}} \\
            \midrule
            ROME & 99.99 &        & 99.49 &        & 91.37 & 74.52 \\
            MEMIT& 99.87 &        & 95.08 &        & 92.36 & 74.20 \\
            \bottomrule
        \end{tabular}
        }
        \label{tab:edit_ent}
        \end{table}

%% file: 6-Conclusion.tex
\section{Conclusion}\label{sec:conclu}
    This paper reveals that relational knowledge in LLMs is encoded not only in MLP layers but also significantly in attention modules. 
    This finding contrasts with previous assumptions that knowledge is primarily stored in MLP weights. 
    Our analysis demonstrates that entity and relational knowledge are stored separately within LLMs, highlighting the complexity of knowledge storage mechanisms. 
    These insights are crucial for improving model interpretability and developing advanced knowledge-based applications. 
    Furthermore, our findings provide a new view for future research and development in LLM-related tasks, such as model editing.

\begin{acks}

The authors of this paper were supported by the National Key R\&D Program of China (No.2022ZD0160503) and the National Natural Science Foundation of China (No.62276264).
\end{acks}

%% file: 0-Main.bbl

\begin{thebibliography}{22}


\ifx \showCODEN    \undefined \def \showCODEN     #1{\unskip}     \fi
\ifx \showDOI      \undefined \def \showDOI       #1{#1}\fi
\ifx \showISBNx    \undefined \def \showISBNx     #1{\unskip}     \fi
\ifx \showISBNxiii \undefined \def \showISBNxiii  #1{\unskip}     \fi
\ifx \showISSN     \undefined \def \showISSN      #1{\unskip}     \fi
\ifx \showLCCN     \undefined \def \showLCCN      #1{\unskip}     \fi
\ifx \shownote     \undefined \def \shownote      #1{#1}          \fi
\ifx \showarticletitle \undefined \def \showarticletitle #1{#1}   \fi
\ifx \showURL      \undefined \def \showURL       {\relax}        \fi
\providecommand\bibfield[2]{#2}
\providecommand\bibinfo[2]{#2}
\providecommand\natexlab[1]{#1}
\providecommand\showeprint[2][]{arXiv:#2}

\bibitem[Chen et~al\mbox{.}(2024a)]%
        {chen2024journey}
\bibfield{author}{\bibinfo{person}{Yuheng Chen}, \bibinfo{person}{Pengfei Cao}, \bibinfo{person}{Yubo Chen}, \bibinfo{person}{Kang Liu}, {and} \bibinfo{person}{Jun Zhao}.} \bibinfo{year}{2024}\natexlab{a}.
\newblock \showarticletitle{Journey to the center of the knowledge neurons: Discoveries of language-independent knowledge neurons and degenerate knowledge neurons}. In \bibinfo{booktitle}{\emph{Proceedings of the AAAI Conference on Artificial Intelligence}}, Vol.~\bibinfo{volume}{38}. \bibinfo{pages}{17817--17825}.
\newblock


\bibitem[Chen et~al\mbox{.}(2024b)]%
        {chen2024knowledge}
\bibfield{author}{\bibinfo{person}{Yuheng Chen}, \bibinfo{person}{Pengfei Cao}, \bibinfo{person}{Yubo Chen}, \bibinfo{person}{Kang Liu}, {and} \bibinfo{person}{Jun Zhao}.} \bibinfo{year}{2024}\natexlab{b}.
\newblock \showarticletitle{Knowledge Localization: Mission Not Accomplished? Enter Query Localization!}
\newblock \bibinfo{journal}{\emph{arXiv preprint arXiv:2405.14117}} (\bibinfo{year}{2024}).
\newblock


\bibitem[Dai et~al\mbox{.}(2022)]%
        {dai2022knowledge}
\bibfield{author}{\bibinfo{person}{Damai Dai}, \bibinfo{person}{Li Dong}, \bibinfo{person}{Yaru Hao}, \bibinfo{person}{Zhifang Sui}, \bibinfo{person}{Baobao Chang}, {and} \bibinfo{person}{Furu Wei}.} \bibinfo{year}{2022}\natexlab{}.
\newblock \showarticletitle{Knowledge Neurons in Pretrained Transformers}. In \bibinfo{booktitle}{\emph{Proceedings of the 60th Annual Meeting of the Association for Computational Linguistics (Volume 1: Long Papers)}}. \bibinfo{pages}{8493--8502}.
\newblock


\bibitem[De~Cao et~al\mbox{.}(2021)]%
        {de2021editing}
\bibfield{author}{\bibinfo{person}{Nicola De~Cao}, \bibinfo{person}{Wilker Aziz}, {and} \bibinfo{person}{Ivan Titov}.} \bibinfo{year}{2021}\natexlab{}.
\newblock \showarticletitle{Editing Factual Knowledge in Language Models}. In \bibinfo{booktitle}{\emph{Proceedings of the 2021 Conference on Empirical Methods in Natural Language Processing}}. \bibinfo{pages}{6491--6506}.
\newblock


\bibitem[Dong et~al\mbox{.}(2022)]%
        {dong2022calibrating}
\bibfield{author}{\bibinfo{person}{Qingxiu Dong}, \bibinfo{person}{Damai Dai}, \bibinfo{person}{Yifan Song}, \bibinfo{person}{Jingjing Xu}, \bibinfo{person}{Zhifang Sui}, {and} \bibinfo{person}{Lei Li}.} \bibinfo{year}{2022}\natexlab{}.
\newblock \showarticletitle{Calibrating Factual Knowledge in Pretrained Language Models}. In \bibinfo{booktitle}{\emph{Findings of the Association for Computational Linguistics: EMNLP 2022}}. \bibinfo{pages}{5937--5947}.
\newblock


\bibitem[Geva et~al\mbox{.}(2021)]%
        {geva2021transformer}
\bibfield{author}{\bibinfo{person}{Mor Geva}, \bibinfo{person}{Roei Schuster}, \bibinfo{person}{Jonathan Berant}, {and} \bibinfo{person}{Omer Levy}.} \bibinfo{year}{2021}\natexlab{}.
\newblock \showarticletitle{Transformer Feed-Forward Layers Are Key-Value Memories}. In \bibinfo{booktitle}{\emph{Proceedings of the 2021 Conference on Empirical Methods in Natural Language Processing}}. \bibinfo{pages}{5484--5495}.
\newblock


\bibitem[Hu et~al\mbox{.}(2024)]%
        {hu2024wilke}
\bibfield{author}{\bibinfo{person}{Chenhui Hu}, \bibinfo{person}{Pengfei Cao}, \bibinfo{person}{Yubo Chen}, \bibinfo{person}{Kang Liu}, {and} \bibinfo{person}{Jun Zhao}.} \bibinfo{year}{2024}\natexlab{}.
\newblock \showarticletitle{Wilke: Wise-layer knowledge editor for lifelong knowledge editing}.
\newblock \bibinfo{journal}{\emph{arXiv preprint arXiv:2402.10987}} (\bibinfo{year}{2024}).
\newblock


\bibitem[Huang et~al\mbox{.}(2022)]%
        {huang2022transformer}
\bibfield{author}{\bibinfo{person}{Zeyu Huang}, \bibinfo{person}{Yikang Shen}, \bibinfo{person}{Xiaofeng Zhang}, \bibinfo{person}{Jie Zhou}, \bibinfo{person}{Wenge Rong}, {and} \bibinfo{person}{Zhang Xiong}.} \bibinfo{year}{2022}\natexlab{}.
\newblock \showarticletitle{Transformer-Patcher: One Mistake Worth One Neuron}. In \bibinfo{booktitle}{\emph{The Eleventh International Conference on Learning Representations}}.
\newblock


\bibitem[Li et~al\mbox{.}(2024)]%
        {li2024relational}
\bibfield{author}{\bibinfo{person}{Pu Li}, \bibinfo{person}{Xiaoyan Yu}, \bibinfo{person}{Hao Peng}, \bibinfo{person}{Yantuan Xian}, \bibinfo{person}{Linqin Wang}, \bibinfo{person}{Li Sun}, \bibinfo{person}{Jingyun Zhang}, {and} \bibinfo{person}{Philip~S Yu}.} \bibinfo{year}{2024}\natexlab{}.
\newblock \showarticletitle{Relational Prompt-based Pre-trained Language Models for Social Event Detection}.
\newblock \bibinfo{journal}{\emph{arXiv preprint arXiv:2404.08263}} (\bibinfo{year}{2024}).
\newblock


\bibitem[Ma et~al\mbox{.}(2023)]%
        {ma2023ex}
\bibfield{author}{\bibinfo{person}{Huanhuan Ma}, \bibinfo{person}{Weizhi Xu}, \bibinfo{person}{Yifan Wei}, \bibinfo{person}{Liuji Chen}, \bibinfo{person}{Liang Wang}, \bibinfo{person}{Qiang Liu}, {and} \bibinfo{person}{Shu Wu}.} \bibinfo{year}{2023}\natexlab{}.
\newblock \showarticletitle{EX-FEVER: A Dataset for Multi-hop Explainable Fact Verification}.
\newblock \bibinfo{journal}{\emph{arXiv preprint arXiv:2310.09754}} (\bibinfo{year}{2023}).
\newblock


\bibitem[Meng et~al\mbox{.}(2022a)]%
        {meng2022locating}
\bibfield{author}{\bibinfo{person}{Kevin Meng}, \bibinfo{person}{David Bau}, \bibinfo{person}{Alex Andonian}, {and} \bibinfo{person}{Yonatan Belinkov}.} \bibinfo{year}{2022}\natexlab{a}.
\newblock \showarticletitle{Locating and editing factual associations in GPT}.
\newblock \bibinfo{journal}{\emph{Advances in Neural Information Processing Systems}}  \bibinfo{volume}{35} (\bibinfo{year}{2022}), \bibinfo{pages}{17359--17372}.
\newblock


\bibitem[Meng et~al\mbox{.}(2022b)]%
        {meng2022mass}
\bibfield{author}{\bibinfo{person}{Kevin Meng}, \bibinfo{person}{Arnab~Sen Sharma}, \bibinfo{person}{Alex~J Andonian}, \bibinfo{person}{Yonatan Belinkov}, {and} \bibinfo{person}{David Bau}.} \bibinfo{year}{2022}\natexlab{b}.
\newblock \showarticletitle{Mass-Editing Memory in a Transformer}. In \bibinfo{booktitle}{\emph{The Eleventh International Conference on Learning Representations}}.
\newblock


\bibitem[Mitchell et~al\mbox{.}(2021)]%
        {mitchell2021fast}
\bibfield{author}{\bibinfo{person}{Eric Mitchell}, \bibinfo{person}{Charles Lin}, \bibinfo{person}{Antoine Bosselut}, \bibinfo{person}{Chelsea Finn}, {and} \bibinfo{person}{Christopher~D Manning}.} \bibinfo{year}{2021}\natexlab{}.
\newblock \showarticletitle{Fast Model Editing at Scale}. In \bibinfo{booktitle}{\emph{International Conference on Learning Representations}}.
\newblock


\bibitem[Mitchell et~al\mbox{.}(2022)]%
        {mitchell2022memory}
\bibfield{author}{\bibinfo{person}{Eric Mitchell}, \bibinfo{person}{Charles Lin}, \bibinfo{person}{Antoine Bosselut}, \bibinfo{person}{Christopher~D Manning}, {and} \bibinfo{person}{Chelsea Finn}.} \bibinfo{year}{2022}\natexlab{}.
\newblock \showarticletitle{Memory-based model editing at scale}. In \bibinfo{booktitle}{\emph{International Conference on Machine Learning}}. PMLR, \bibinfo{pages}{15817--15831}.
\newblock


\bibitem[Petroni et~al\mbox{.}(2019)]%
        {petroni2019language}
\bibfield{author}{\bibinfo{person}{Fabio Petroni}, \bibinfo{person}{Tim Rockt{\"a}schel}, \bibinfo{person}{Sebastian Riedel}, \bibinfo{person}{Patrick Lewis}, \bibinfo{person}{Anton Bakhtin}, \bibinfo{person}{Yuxiang Wu}, {and} \bibinfo{person}{Alexander Miller}.} \bibinfo{year}{2019}\natexlab{}.
\newblock \showarticletitle{Language Models as Knowledge Bases?}. In \bibinfo{booktitle}{\emph{Proceedings of the 2019 Conference on Empirical Methods in Natural Language Processing and the 9th International Joint Conference on Natural Language Processing (EMNLP-IJCNLP)}}. \bibinfo{pages}{2463--2473}.
\newblock


\bibitem[Sinitsin et~al\mbox{.}(2019)]%
        {sinitsin2019editable}
\bibfield{author}{\bibinfo{person}{Anton Sinitsin}, \bibinfo{person}{Vsevolod Plokhotnyuk}, \bibinfo{person}{Dmitry Pyrkin}, \bibinfo{person}{Sergei Popov}, {and} \bibinfo{person}{Artem Babenko}.} \bibinfo{year}{2019}\natexlab{}.
\newblock \showarticletitle{Editable Neural Networks}. In \bibinfo{booktitle}{\emph{International Conference on Learning Representations}}.
\newblock


\bibitem[Wei et~al\mbox{.}(2023)]%
        {wei2023menatqa}
\bibfield{author}{\bibinfo{person}{Yifan Wei}, \bibinfo{person}{Yisong Su}, \bibinfo{person}{Huanhuan Ma}, \bibinfo{person}{Xiaoyan Yu}, \bibinfo{person}{Fangyu Lei}, \bibinfo{person}{Yuanzhe Zhang}, \bibinfo{person}{Jun Zhao}, {and} \bibinfo{person}{Kang Liu}.} \bibinfo{year}{2023}\natexlab{}.
\newblock \showarticletitle{MenatQA: A New Dataset for Testing the Temporal Comprehension and Reasoning Abilities of Large Language Models}. In \bibinfo{booktitle}{\emph{Findings of the Association for Computational Linguistics: EMNLP 2023}}. \bibinfo{pages}{1434--1447}.
\newblock


\bibitem[Xia et~al\mbox{.}(2022)]%
        {xia2022medconqa}
\bibfield{author}{\bibinfo{person}{Fei Xia}, \bibinfo{person}{Bin Li}, \bibinfo{person}{Yixuan Weng}, \bibinfo{person}{Shizhu He}, \bibinfo{person}{Kang Liu}, \bibinfo{person}{Bin Sun}, \bibinfo{person}{Shutao Li}, {and} \bibinfo{person}{Jun Zhao}.} \bibinfo{year}{2022}\natexlab{}.
\newblock \showarticletitle{MedConQA: medical conversational question answering system based on knowledge graphs}. In \bibinfo{booktitle}{\emph{Proceedings of the 2022 Conference on Empirical Methods in Natural Language Processing: System Demonstrations}}. \bibinfo{pages}{148--158}.
\newblock


\bibitem[Yu et~al\mbox{.}(2021)]%
        {yu2021multi}
\bibfield{author}{\bibinfo{person}{Xiaoyan Yu}, \bibinfo{person}{Qingbin Liu}, \bibinfo{person}{Shizhu He}, \bibinfo{person}{Kang Liu}, \bibinfo{person}{Shengping Liu}, \bibinfo{person}{Jun Zhao}, {and} \bibinfo{person}{Yongbin Zhou}.} \bibinfo{year}{2021}\natexlab{}.
\newblock \showarticletitle{Multi-strategy knowledge distillation based teacher-student framework for machine reading comprehension}. In \bibinfo{booktitle}{\emph{China National Conference on Chinese Computational Linguistics}}. Springer, \bibinfo{pages}{209--225}.
\newblock


\bibitem[Yu et~al\mbox{.}(2024)]%
        {yu2024neeko}
\bibfield{author}{\bibinfo{person}{Xiaoyan Yu}, \bibinfo{person}{Tongxu Luo}, \bibinfo{person}{Yifan Wei}, \bibinfo{person}{Fangyu Lei}, \bibinfo{person}{Yiming Huang}, \bibinfo{person}{Peng Hao}, {and} \bibinfo{person}{Liehuang Zhu}.} \bibinfo{year}{2024}\natexlab{}.
\newblock \showarticletitle{Neeko: Leveraging Dynamic LoRA for Efficient Multi-Character Role-Playing Agent}.
\newblock \bibinfo{journal}{\emph{arXiv preprint arXiv:2402.13717}} (\bibinfo{year}{2024}).
\newblock


\bibitem[Zheng et~al\mbox{.}(2023)]%
        {zheng2023can}
\bibfield{author}{\bibinfo{person}{Ce Zheng}, \bibinfo{person}{Lei Li}, \bibinfo{person}{Qingxiu Dong}, \bibinfo{person}{Yuxuan Fan}, \bibinfo{person}{Zhiyong Wu}, \bibinfo{person}{Jingjing Xu}, {and} \bibinfo{person}{Baobao Chang}.} \bibinfo{year}{2023}\natexlab{}.
\newblock \showarticletitle{Can We Edit Factual Knowledge by In-Context Learning?}. In \bibinfo{booktitle}{\emph{Proceedings of the 2023 Conference on Empirical Methods in Natural Language Processing}}. \bibinfo{pages}{4862--4876}.
\newblock


\bibitem[Zhong et~al\mbox{.}(2023)]%
        {zhong2023mquake}
\bibfield{author}{\bibinfo{person}{Zexuan Zhong}, \bibinfo{person}{Zhengxuan Wu}, \bibinfo{person}{Christopher~D Manning}, \bibinfo{person}{Christopher Potts}, {and} \bibinfo{person}{Danqi Chen}.} \bibinfo{year}{2023}\natexlab{}.
\newblock \showarticletitle{MQuAKE: Assessing Knowledge Editing in Language Models via Multi-Hop Questions}. In \bibinfo{booktitle}{\emph{Proceedings of the 2023 Conference on Empirical Methods in Natural Language Processing}}. \bibinfo{pages}{15686--15702}.
\newblock


\end{thebibliography}
